\documentclass[10pt,twocolumn,letterpaper]{article}

\usepackage{wacv}
\usepackage{times}
\usepackage{epsfig}
\usepackage{graphicx}
\usepackage{amsmath}
\usepackage{amssymb}

\newcommand{\mat}[1]{\mathbf{#1}}
\newcommand{\norm}[1]{\left|\left| #1 \right| \right|}
\newcommand{\abs}[1]{\left| #1 \right|}
\newcommand{\R}{\mathbb{R}}
\usepackage{multirow}
\usepackage{hhline}
\usepackage{mathtools}
\usepackage{caption, subcaption}

\DeclarePairedDelimiter\floor{\lfloor}{\rfloor}
\DeclarePairedDelimiter\inner{\langle}{\rangle}
\DeclareMathOperator*{\argmax}{argmax} 
\newcommand{\myparagraph}[1]{\smallskip\noindent\textbf{#1.}}

\usepackage[pagebackref=true,breaklinks=true,letterpaper=true,colorlinks,bookmarks=false]{hyperref}

\wacvfinalcopy 

\def\wacvPaperID{133} 

\ifwacvfinal\fi
\begin{document}

\title{End-to-End CRF and Discriminative Dictionary Learning \\ for Semantic Segmentation of Fine-Grained Actions}
\title{Semantic Segmentation of Fine-Grained Actions Via \\ End-to-End CRF and Discriminative Dictionary Learning}
\title{End-to-End Fine-Grained Action Segmentation and Recognition \\Via Joint CRF and Dictionary Learning}
\title{End-to-End Fine-Grained Action Segmentation and Recognition \\Via Joint CRF and Discriminative Sparse Coding}
\title{End-to-End Semantic Segmentation of Fine-Grained Actions \\Using Conditional Random Field Models and Discriminative Sparse Coding}
\title{End-to-End Fine-Grained Action Segmentation and Recognition Using Conditional Random Field Models and Discriminative Sparse Coding}

\author{Effrosyni Mavroudi$^\dag$, Divya Bhaskara$^{\dag\ddag}$, Shahin Sefati$^{\dag\natural}$, Haider Ali$^\dag$ and Ren\'e Vidal$^\dag$\\
$^\dag$Johns Hopkins University,
$^\ddag$University of Virginia,
$^\natural$Comcast AI Research\\
}

\maketitle

\begin{abstract}
Fine-grained action segmentation and recognition is an important yet challenging task. Given a long, untrimmed sequence of kinematic data, the task is to classify the action at each time frame and segment the time series into the correct sequence of actions. In this paper, we propose a novel framework that combines a temporal Conditional Random Field (CRF) model with a powerful frame-level representation based on discriminative sparse coding. We introduce an end-to-end algorithm for jointly learning the weights of the CRF model, which include action classification and action transition costs, as well as an overcomplete dictionary of mid-level action primitives. This results in a CRF model that is driven by sparse coding features obtained using a discriminative dictionary that is shared among different actions and adapted to the task of structured output learning. We evaluate our method on three surgical tasks using kinematic data from the JIGSAWS dataset, as well as on a food preparation task using accelerometer data from the 50 Salads dataset. Our results show that the proposed method performs on par or better than state-of-the-art methods.
\end{abstract}

\section{Introduction}

Temporal segmentation and recognition of complex activities in long continuous recordings is a useful, yet challenging task. Examples of complex activities comprised of fine-grained goal-driven actions that follow a grammar are surgical procedures~\cite{Gao2014}, food preparation~\cite{Stein:UbiComp13} and assembly tasks~\cite{Vo:CVPR14}. For instance, in the medical field there is a need to better train surgeons in performing surgical procedures using new technologies such as the daVinci robot. One possible approach is to use machine learning and computer vision techniques to automatically determine the skill level of the surgeon from kinematic data of the surgeon's performance recorded by the robot~\cite{Gao2014}. Such an approach typically requires an accurate classification of the surgical gesture at each time frame \cite{Bejar:MICCAI12} and a segmentation of the surgical task into the correct sequence of gestures \cite{Tao:MICCAI13}. Another example of a complex activity with goal-driven fine-grained actions following a grammar is cooking. Although the actions performed while preparing a recipe and their relative ordering can vary, there are still temporal relations among them. For instance, the action \emph{stir milk} usually happens after \emph{pour milk}, or the action \emph{fry egg} usually follows the action \emph{crack egg}. Robots equipped with the ability to automatically recognize actions during food preparation could assist individuals with cognitive impairments in their daily activities by providing prompts and instructions. However, the task of fine-grained action segmentation and recognition is challenging due to the subtle differences between actions, the variability in the duration and style of execution among users and the variability in the relative ordering of actions.

Existing approaches to fine-grained action segmentation and recognition use a \emph{temporal model} to capture the temporal evolution and ordering of actions, such as Hidden Markov Models (HMMs)~\cite{Kuehne:CVPR14,Tao:IPCAI12}, Conditional Random Fields (CRF)~\cite{Lea:WACV15,Lea:ICRA16}, Markov semi-Markov Conditional Random Fields (MsM-CRF)~\cite{Tao:MICCAI13}, Recurrent Neural Networks~\cite{Dipietro:MICCAI16,Richard:CoRR17} and Temporal Convolutional Networks (TCNs)~\cite{Lea:CVPR17}. However, such models cannot capture subtle differences between actions without a powerful, discriminative and robust \emph{representation of frames} or short temporal segments. Sparse coding has emerged as a powerful signal representation in which the raw data in a certain time frame is represented as a linear combination of a small number of basis elements from an overcomplete dictionary. The coefficients of this linear combination are called \emph{sparse codes} and are used as a new representation for temporal modeling. However, since the dictionary is typically learned in an unsupervised manner by minimizing a regularized reconstruction error \cite{Elad:TSP06}, the resulting representation may not be discriminative for a given learning task. Task-driven discriminative dictionary learning addresses this issue by coupling dictionary and classifier learning~\cite{Mairal:NIPS09}. For example, Sefati et al.~\cite{Sefati:M2CAI15} propose an approach to fine-grained action recognition called Shared Discriminative Sparse Dictionary Learning (SDSDL), where sparse codes are extracted at each time frame and a frame feature is computed by average pooling the sparse codes over a short temporal window surrounding the frame. The dictionary is jointly learned with the per-frame classifier parameters, resulting in a discriminative mid-level representation that is shared across all actions/gestures. However, their approach lacks a temporal model, which is crucial for modeling temporal dependencies. Although prior work \cite{Yang:PAMI17}  has combined discriminative dictionary learning with CRFs for the purpose of saliency detection, such work is not directly applicable to fine-grained action recognition.

In this work we propose a joint model for fine-grained action recognition and segmentation that integrates a CRF for temporal modeling and discriminative sparse coding for frame-wise action representation. The proposed CRF models the temporal structure of long untrimmed activities via unary potentials that represent the cost of assigning an action label to a frame-wise representation of an action obtained via discriminative sparse coding, and pairwise potentials that capture the transitions between actions and encourage smoothness of the predicted label sequence. The parameters of the combined model are trained jointly in an end-to-end manner using a max-margin approach. Our experiments show competitive performance in the task of fine-grained action recognition, especially in the regime of limited training data. In summary, the contributions of this paper are three-fold:
\begin{enumerate}
\item We propose a novel framework for fine-grained action segmentation and recognition which uses a CRF model whose target variables (action labels per time step) are conditioned on sparse codes.
\item We introduce an algorithm for training our model in an end-to-end fashion. In particular, we jointly learn a task-specific discriminative dictionary and the CRF unary and pairwise weights by using Stochastic Gradient Descent (SGD).
\item We evaluate our model on two public datasets focused on goal-driven complex activities comprised of fine-grained actions. In particular, we use robot kinematic data from the JHU-ISI Gesture and Skill Assessment Working Set (JIGSAWS)~\cite{Gao2014} dataset and evaluate our method on three surgical tasks. We also experiment with accelerometer data from the 50 Salads~\cite{Stein:UbiComp13} dataset for recognizing actions that are labeled at two levels of granularity. Results show that our method performs on par with most state-of-the-art methods.
\end{enumerate}
\section{Related Work}
The task of fine-grained action segmentation and recognition has recently received increased attention due to the release of datasets such as MPII Cooking~\cite{Rohrbach:CVPR12}, JIGSAWS~\cite{Gao2014} and 50 Salads~\cite{Stein:UbiComp13}. In this section, we briefly review some of the main existing approaches for tackling this problem. Besides, we briefly discuss existing work on discriminative dictionary learning. Note that since the focus of this paper is fine-grained action recognition from kinematic data, we do not discuss approaches for feature extraction or object parsing from video data.

\myparagraph{Fine-grained action recognition from kinematic data}
A straightforward approach to action segmentation and classification is the use of overlapping temporal windows in conjunction with temporal segment classifiers and non-maximum suppression (e.g.,~\cite{Rohrbach:CVPR12, Oneata:ICCV13}). However this approach does not exploit long-range temporal dependencies.

Recently, deep learning approaches have started to emerge in the field. For instance, in ~\cite{Dipietro:MICCAI16} a recurrent neural network (Long Short Term Memory network - LSTM) is applied to kinematic data, while in~\cite{Lea:CVPR17} a Temporal Convolutional Network composed of 1D convolutions, non-linearities and pooling/upsampling layers is introduced. Although these models yield promising results, they do not explicitly model correlations and dependencies among action labels.

Another line of work, including our proposed method, takes into account the fact that the action segmentation and classification problem is a structured output prediction problem due to the temporal structure of the sequence of action labels and thus employs structured temporal models  such as HMMs and their extensions~\cite{Tao:IPCAI12,Kuehne:CVPR14,Kuehne:WACV16}. Among them, the work that is most related to this work is  Sparse-HMMs~\cite{Tao:IPCAI12}, which combines dictionary learning with HMMs. However, a Sparse-HMM is a generative model in which a separate dictionary is learned for each action class. In this work we use a CRF, which is a discriminative model, and we learn a dictionary that is shared among all action classes. Discriminative models like CRFs~\cite{Lea:WACV15,Lea:ICRA16}, semi-Markov CRFs~\cite{Tao:MICCAI13} have gained popularity since they allow for flexible energy functions. Other types of temporal models include a duration model and language model recently proposed in~\cite{Richard:CVPR16} for modeling action durations and context. The input to these temporal models are either the kinematic data themselves or features extracted from them. For instance, in the Latent Convolutional Skip Chain CRF (LC-SC-CRF)~\cite{Lea:ICRA16} the responses to convolutional filters, which capture latent action primitives, are used as features. 

\myparagraph{Discriminative Dictionary Learning}
Task-driven discriminative dictionary learning was introduced in the seminal work of Mairal et al.~\cite{Mairal:NIPS09} and couples the process of dictionary learning and classifier training, thus incorporating supervised learning to sparse coding. Since then discriminative dictionary learning has enjoyed many successes in diverse areas such as handwritten digit classification~\cite{Mairal:PAMI12,Yang:CVPR10-supervised}, face recognition~\cite{Jiang:CVPR11,Yang:CVPR10-supervised,Quan:CVPR16}, object category recognition~\cite{Jiang:CVPR11,Quan:CVPR16,Boureau:CVPR10}, scene classification~\cite{Boureau:CVPR10,Lian:ECCV10,Quan:CVPR16}, and action classification~\cite{Quan:CVPR16}.

The closest work to ours is the Shared Discriminative Sparse Dictionary Learning (SDSDL) proposed by Sefati et al.~\cite{Sefati:M2CAI15}, where sparse codes are used as frame features and a discriminative dictionary is jointly learned with per frame action classifiers for the task of surgical task segmentation. Our work builds on top of this model by replacing the per-frame classifiers, which compute independent predictions per frame, with a structured output temporal model (CRF), which takes into account the temporal dependencies between actions. While prior work has considered joint dictionary and CRF learning~\cite{Tao:ECCV14,Yang:CVPR12,Yang:PAMI17} for the tasks of semantic segmentation and saliency estimation, our work differs from these previous approaches in three key aspects. First, to the best of our knowledge, we are the first to apply joint dictionary and CRF learning to the task of action segmentation and classification.
Second, we are learning unary CRF classifiers and pairwise transition scores, while in~\cite{Tao:ECCV14} only two scalar variables encoding the relative weight between the unary and pairwise potentials are learned. Third, we use local temporal average-pooling of sparse codes as a feature extraction process for capturing local temporal context instead of the raw sparse codes used in ~\cite{Yang:CVPR12,Yang:PAMI17}. 

\section{Technical Approach}

\begin{figure}
\centering
\def\svgscale{0.5}
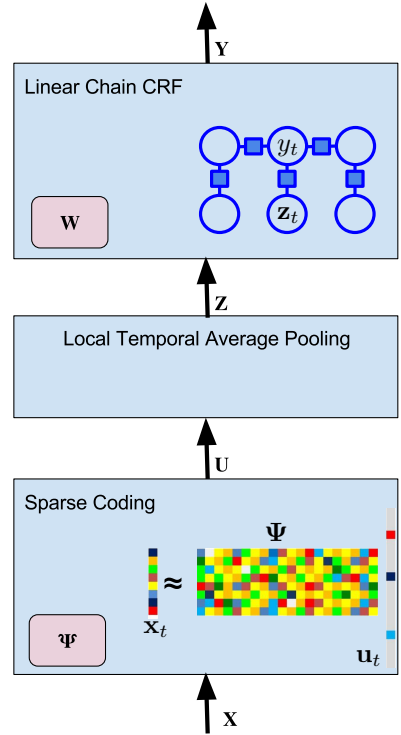
\caption{Overview of our framework. Given an input time series $\mat{X}$, we first extract sparse codes $\mat{U}$ for each timestep using a dictionary $\boldsymbol{\Psi}$. Sparse codes are then average pooled in short temporal windows yielding feature vectors $\mat{Z}$ per timestep. These feature vectors are then given as inputs to a Linear Chain CRF with weights $\mat{W}$. Trainable parameters $\boldsymbol{\Psi}$ and $\mat{W}$ are shown in light pink boxes.}
\label{fig:overview}
\end{figure}

In this section, we introduce our temporal CRF model and frame-wise representation based on sparse coding and describe our algorithm for training our model. Figure \ref{fig:overview} illustrates the key components of our model.

\subsection{Model}
\myparagraph{Frame-wise representation}
Let $\mat{X} = \{ \mat{x}_t \}_{t=1}^T$ be a sequence of length $T$, with $\mat{x}_t \in \R^p$ being the input at time $t$ (e.g., the robot's joint positions and velocities). Our goal is to compactly represent each $\mat{x}_t$ as a linear combination of a small number of atomic motions using an overcomplete dictionary of representative atomic motions $\boldsymbol{\Psi} \in \R^{p \times m}$, i.e., $\mat{x}_t \approx \boldsymbol{\Psi} \mat{u}_t$, where $\mat{u}_t \in \R^m$ is the vector of sparse coefficients obtained for frame $t$. Such sparse codes can be obtained by considering the following optimization problem:
\begin{equation}
\min_{\{\mat{u}_t\}_{t=1}^T} \frac{1}{T} \displaystyle \sum_{t=1}^T \norm{\mat{x}_t - \boldsymbol{\Psi} \mat{u}_t}_2^2 + \lambda_u \norm{\mat{u}_t}_1,
\label{eq:sparse_coding}
\end{equation}
where $\lambda_u$ is a regularization parameter controlling the trade-off between reconstruction error and sparsity of the coefficients. Problem \eqref{eq:sparse_coding} is a standard Lasso regression and can be efficiently solved using existing sparse coding algorithms~\cite{Mairal:JMLR2010}. After computing sparse codes $\mat{u}_t$ for each time step of the input sequence, we follow the approach proposed in~\cite{Sefati:M2CAI15} to compute feature vectors $\{ \mat{z}_t \}_{t=1}^T$. Namely, we initially split the positive and negative components of the sparse codes and stack them on top of each other. This step yields a vector $\mat{a}_t \in \R^D$, $D = 2m$, which is given by: 
\begin{equation}
\mat{a}_t = 
\begin{bmatrix}
\max(0, \mat{u}_t) \\
\min(0, \mat{u}_t)
\end{bmatrix}.
\label{eq:alphas}
\end{equation}
This is a common practice~\cite{Coates:ICML11-sparse,Bo:CVPR13}, which allows the classification layer to assign different weights to positive and negative responses. Second, we compute a feature vector $\mat{z}_t \in \R^D$ for each frame by average-pooling vectors $\mat{a}_t$ in a temporal window $T_t$ surrounding frame $t$, i.e.:
\begin{equation}
\mat{z}_t = \frac{1}{L} \displaystyle \sum_{j \in
T_t } \mat{a}_j, \quad 
T_t \doteq 
 \left\{t-\floor*{\frac{L}{2}}, t+\floor*{\frac{L}{2}}\right\},
\label{eq:zs}
\end{equation}
where $L$ is the length of the temporal window centered at frame $t$. This feature vector captures local temporal context.

\myparagraph{Temporal model} Let $\mat{Z} = \{ \mat{z}_t\}_{t=1}^T$ be a sequence of length $T$ with $\mat{z}_t$ being the feature vector representing the input at time $t$, and $\mat{Y} = \{ y_t\}_{t=1}^T$ be the corresponding sequence of action labels per frame, $y_t \in \{ 1, \ldots, N_c\}$, with $N_c$ being the number of action classes. Let $G = \{ \cal{V}, \cal{E}\}$ be the graph whose nodes correspond to different frames ($\abs{\cal{V}}=T$) and whose edges connect every $d$ frames (with $d=1$ corresponding to consecutive frames). Our CRF models the conditional distribution of labels given the input features with a Gibbs distribution of the form $P( \mat{Y} \mid \mat{Z}) \propto \exp{E(\mat{Z},\mat{Y})}$, where the energy $E(\mat{Z}, \mat{Y})$ is factorized into a sum of potential functions defined on cliques of order less than or equal to two. Formally, the energy function can be written as:
\begin{equation}
E(\mat{Z}, \mat{Y}) = \displaystyle \sum_{t=1}^T \mat{U}^\top_{y_t} \mat{z}_t + \sum_{t=1}^{T-d} \mat{P}_{y_t, y_{t+d}},
\label{eq:un_pw_energy}
\end{equation}
where the first term is the unary potential which models the score of assigning label $y_t$ to frame $t$ described by feature $\mat{z}_t$, while the second term is called pairwise potential and models the score of assigning labels $y_t$ and $y_{t+d}$ to frames $t$ and $t+d$ respectively ($d$ is a parameter called the skip length and a CRF with $d>1$ is called Skip-Chain CRF (SC-CRF)~\cite{Lea:WACV15,Lea:ICRA16}). $\mat{U}_{y_t} \in \R^D$ is a linear unary classifier corresponding to action class $y_t$ and $\mat{P} \in \R^{N_c \times N_c}$ is the pairwise transition matrix. Note that there exist different variants to this model. For instance, one can use precomputed unary and pairwise potentials and learn two scalar coefficients that encode the relative weights of the two terms~\cite{Tao:ECCV14}. 

We now show how this energy can be written as a linear function with respect to a parameter vector $\mat{W} \in \R^{N_cD + N_c^2}$. The unary term can be rewritten as follows:
\begin{align}
\begin{split}
\sum_{t=1}^T \mat{U}^T_{y_t} \mat{z}_t 
& = \left[ \mat{U}_1^\top, \ldots, \mat{U}_{N_c}^\top \right] 
\begin{bmatrix}
\displaystyle \sum_{t=1}^T \mat{z}_t \delta(y_t = 1) \\
\vdots \\
\displaystyle \sum_{t=1}^T \mat{z}_t \delta(y_t = N_c) \\
\end{bmatrix} \\
& = \mat{W_U}^\top \boldsymbol{\Phi_U}(\mat{Z}, \mat{Y}),
\end{split}
\label{eq:unary_joint_feat}
\end{align}
where $\mat{W_U}$ and $\boldsymbol{\Phi_U}(\mat{Z}, \mat{Y}) \in \R^{N_c D}$ are, respectively, the unary CRF weights and the unary joint feature. Similarly the pairwise term can be written as:
\begin{align}
\begin{split}
&\left[ \mat{P}_{11}, 
\ldots, \mat{P}_{N_c N_c}\right] 
\begin{bmatrix}
\displaystyle \sum_{t=1}^{T-d} \delta(y_t = 1)\delta(y_{t+d}=1) \\
\vdots \\
\displaystyle \sum_{t=1}^{T-d} \delta(y_t = N_c)\delta(y_{t+d}=N_c) \\
\end{bmatrix}  \\
& = \mat{W_P}^\top \boldsymbol{\Phi_P}(\mat{Y}),
\end{split}
\label{eq:pw_joint_feat}
\end{align}
where $\mat{W_P}, \boldsymbol{\Phi_P}(\mat{Y}) \in \R^{N_c^2}$ are the pairwise CRF weights and pairwise joint feature. Therefore, the overall energy function can be written as:
\begin{equation}
E(\mat{Z}, \mat{Y}) = 
\begin{bmatrix}
\mat{W_U} \\
\mat{W_P}
\end{bmatrix}^\top 
\begin{bmatrix}
\boldsymbol{\Phi_U}(\mat{Z}, \mat{Y}) \\
\boldsymbol{\Phi_P}(\mat{Y})
\end{bmatrix} 
= \mat{W}^\top \boldsymbol{\Phi}(\mat{Z}, \mat{Y}),
\label{eq:ssvm_energy}
\end{equation}
where $\mat{W}$ is the vector of CRF weights and $\boldsymbol{\Phi}(\mat{Z}, \mat{Y})$ the joint feature~\cite{Joachims:JMLR09}. At this point, we should emphasize that feature vectors $\mat{Z} = \left[\mat{z}_1, \ldots, \mat{z}_T \right]$ are constructed by local average pooling of the sparse codes and are therefore implicitly dependent of the input data $\mat{X}$ and the dictionary $\boldsymbol{\Psi}$. For the rest of this manuscript, we will denote this dependency by substituting $\mat{Z}$ with the notation $\mat{Z}(\mat{X}, \boldsymbol{\Psi})$. So our energy can be rewritten as: 
\begin{equation}
E(\mat{Z}(\mat{X}, \boldsymbol{\Psi}), \mat{Y})
= \mat{W}^\top \boldsymbol{\Phi}(\mat{Z}(\mat{X}, \boldsymbol{\Psi}), \mat{Y}).
\label{eq:ssvm_energy_nonlinear}
\end{equation}
It should be now clear that if $\boldsymbol{\Psi}$ is fixed, then the energy is linear with respect to the parameter vector $\mat{W}$, like in a standard CRF model. However, if $\boldsymbol{\Psi}$ is a parameter that needs to be learned, then the energy function is nonlinear with respect to $(\mat{W},\boldsymbol{\Psi})$ and thus training is not straightforward. The training problem is addressed next.

\subsection{Training}
Let $\{ \mat{X}^n \}_{n=1}^{N_s}$ be $N_s$ training sequences with associated label sequences $\{\mat{Y}^n\}_{n=1}^{N_s}$. We formulate the training problem as one of minimizing the following regularized loss:
\begin{align}
& J(\mat{W}, \boldsymbol{\Psi}) = \frac{1}{2} \norm{\mat{W}}_F^2 + \nonumber \\
& + \frac{C}{N_s} \displaystyle \sum_{n=1}^{N_s}  \max_{\mat{Y}} \left[
\Delta(\mat{Y}^n, \mat{Y}) + \inner*{\mat{W}, \boldsymbol{\Phi}(\mat{Z}^n(\mat{X}^n, \boldsymbol{\Psi}), \mat{Y})} \right]  \nonumber\\
& - \inner*{\mat{W}, \boldsymbol{\Phi}(\mat{Z}^n(\mat{X}^n, \boldsymbol{\Psi}), \mat{Y}^n)},
\label{eq:max_margin_loss}
\end{align}
where $C$ is a regularization parameter controlling the regularization of the CRF weights, $\Delta(\hat{\mat{Y}}, \mat{Y}) = \displaystyle \sum_{t=1}^T \delta(\hat{y}_t \neq y_t)$ is the Hamming loss between two sequences of labels $\hat{\mat{Y}}$ and $\mat{Y}$, and $\mat{Z}^n$ is the matrix of feature vectors extracted from the frames of input sequence $\mat{X}^n$, i.e., $\mat{Z}^n = \left[ \mat{z}_1^n, \ldots, \mat{z}_T^n \right]$. This max-margin formulation performs regularized empirical risk minimization and bounds the hamming loss from above. We use a Stochastic Gradient Descent algorithm for minimizing the objective function in Eq.~\eqref{eq:max_margin_loss}. Our algorithm is based on the task-driven dictionary learning approach developed by Mairal et al.~\cite{Mairal:NIPS09}. 
Notice that, although the sparse coefficients are computed by minimizing a non-differentiable objective function (Eq.~\ref{eq:sparse_coding}), $J(\mat{W}, \boldsymbol{\Psi})$ is differentiable and its gradient can be computed~\cite{Mairal:PAMI12}. In particular, the function relating the sparse codes $\mat{u}_t$ and the dictionary is differentiable almost everywhere, except at the points where the set of non-zero elements of $\mat{u}_t$ (called the support set and denoted by $S_t$) changes. Assuming that the perturbations of the dictionary atoms are small so that the support set stays the same, we can compute the gradient of the non-zero coefficients with respect to the columns of $\boldsymbol{\Psi}$ indexed by $S_t$, denoted as $\boldsymbol{\Psi}_{S_t}$, as follows~\cite{Tao:ECCV14}:
\begin{equation}
\frac{\partial \mat{u}_t(k)}{\partial \boldsymbol{\Psi}_{S_t}} = (\mat{x}_t - \boldsymbol{\Psi}_{S_t} (\mat{u}_t)_{S_t})(\mat{A}_t^{-1})_{[k]} - (\boldsymbol{\Psi}_{S_t} \mat{A}_t^{-\top})_{\inner*{k}} (\mat{u}_t)_{S_t}^\top
\label{eq:u_diff}
\end{equation}
where $k \in S_t$, $(\mat{u}_t)_{S_t}$ denotes the sub-vector of $\mat{u}_t$ with entries in $S_t$, $\mat{A}_t = \boldsymbol{\Psi}_{S_t}^\top \boldsymbol{\Psi}_{S_t}$, and the subscripts $[k]$ and $\inner{k}$ denote, respectively, the $k$-th row and column of the corresponding matrix. 

Given the dictionary and CRF weights computed at the $(i-1)$-th iteration, the main s-eps-converted-to.pdf of our iterative algorithm at the $i$-th iteration are:
\begin{enumerate}
\item Randomly select a training sequence $(\mat{X}^i, \mat{Y}^i)$.
\item Compute sparse codes $\mat{u}_t$ with Eq.~\ref{eq:sparse_coding} and feature vectors $\mat{z}_t$ with Eq.~\ref{eq:zs} using dictionary $\boldsymbol{\Psi}^{(i-1)}$. 
\item Find the sequence $\hat{\mat{Y}}$ that yields the most violated constraint by solving the loss augmented inference problem:
\begin{align}
\hat{\mat{Y}} &= \argmax_{\mat{Y}} 
\Delta(\mat{Y}^i, \mat{Y}) + \nonumber\\
&\inner*{\mat{W}^{(i-1)}, \boldsymbol{\Phi}(\mat{Z}^i(\mat{X}^i, \boldsymbol{\Psi}^{(i-1)}), \mat{Y})} 
\label{eq:loss_aug_inference}
\end{align}
using the Viterbi algorithm (see~\cite{Lea:ICRA16} for details regarding inference when using a SC-CRF ($d > 1$)).
\item Compute gradient with respect to the CRF parameters $W$:
\begin{align}
\frac{\partial J}{\partial \mat{W}} &= \mat{W}^{(i-1)} + C(\boldsymbol{\Phi}(\mat{Z}^i(\mat{X}^i, \boldsymbol{\Psi}^{(i-1)}), \hat{\mat{Y}}) -
\nonumber\\
&- \boldsymbol{\Phi}(\mat{Z}^i(\mat{X}^i, \boldsymbol{\Psi}^{(i-1)}), \mat{Y}^i)).
\label{eq:label_w}
\end{align}

\item Compute gradients with respect to the dictionary $\boldsymbol{\Psi}$ using the chain rule:
\begin{align}
& \frac{\partial J}{\partial \boldsymbol{\Psi}} = \displaystyle \sum_{t=1}^T \left(\frac{\partial J}{\partial \mat{z}_t}\right)^\top \frac{\partial \mat{z}_t}{\partial \boldsymbol{\Psi}} 
= \displaystyle \sum_{t=1}^T \left(\frac{\partial J}{\partial \mat{z}_t}\right)^\top \frac{1}{L} \sum_{j\in T_t} \frac{\partial \mat{a}_j}{\partial \boldsymbol{\Psi}} , \quad 
\nonumber\\
& = \frac{1}{L} \displaystyle \sum_{t=1}^T \sum_{j\in T_t} (\mat{x}_j - \boldsymbol{\Psi}^{(i-1)}_{S_j} (\mat{u}_j)_{S_j})\left(\mat{A}_j^{-1} \left( \frac{\partial J}{\partial \mat{z}_t}\right)_{\tilde{S}_j}\right)^\top \nonumber\\
& - \boldsymbol{\Psi}^{(i-1)}_{S_j} \mat{A}_j^{-\top} (\mat{u}_j)_{S_j} \left( \frac{\partial J}{\partial \mat{z}_t}\right)_{\tilde{S}_j}^\top ,
\label{eq:grad_psi}
\end{align}
where $\frac{\partial J}{\partial \mat{z}_t} = \mat{U}_{\hat{y}_t} - \mat{U}_{y_t} \in \R^D$, $S_j$ is the set of indices corresponding to the non-zero entries of the vector $\mat{u}_j$, $\tilde{S}_j$ is the set of indices corresponding to the non-zero entries of the vector $\mat{a}_j$, $A_j = \boldsymbol{\Psi}_{S_j}^\top \boldsymbol{\Psi}_{S_j}$, $\boldsymbol{\Psi}_{S_j}$ denotes the active columns of the dictionary indexed by $S_j$, $(\mat{u}_j)_{S_j}$ denotes the non-zero entries of vector $\mat{u}_j$ and $(\frac{\partial J}{\partial \mat{z}_t})_{\tilde{S}_j}$ denotes the entries of the partial derivative corresponding to non-zero entries of vector $\mat{a}_j$.
\item Update $\mat{W}$, $\boldsymbol{\Psi}$ using stochastic gradient descent.

\item Normalize the dictionary atoms to have unit $l_2$ norm.
This step prevents the columns of $\boldsymbol{\Psi}$ from becoming arbitrarily large, which would result in arbitrarily small sparse coefficients. 
\end{enumerate}
\section{Experiments}
\begin{table*}[ht!]
\setlength{\tabcolsep}{0.75mm}
\centering
\begin{tabular}{|@{\;}l@{\;}|ccc|ccc|}
\hline
Method & \multicolumn{3}{|c|}{LOSO} & \multicolumn{3}{|c|}{LOUO} \\
\hline
 & SU & KT & NP & SU & KT & NP\\
\hline
GMM-HMM~\cite{Ahmidi:TBME17}  &82.22  & 80.95  & 70.55 &  73.95  & 72.47  & 64.13 \\
KSVD-SHMM~\cite{Tao:IPCAI12,Ahmidi:TBME17} & 83.40 & 83.54  & 73.09  & 73.45  & 74.89   & 62.78  \\
MsM-CRF~\cite{Tao:MICCAI13,Ahmidi:TBME17} & 81.99  & 79.26  & 72.44  & 67.84   & 44.68  & 63.28 \\
SC-CRF-SL~\cite{Lea:WACV15,Ahmidi:TBME17} & 85.18 & \textbf{84.03} & \emph{75.09} & 81.74 & \textbf{78.95} & \textbf{74.77} \\
SDSDL~\cite{Sefati:M2CAI15} & \textbf{86.32} & 82.54 & 74.88 & 78.68 & 75.11 & 66.01 \\
LSTM (5Hz)~\cite{Dipietro:MICCAI16}\text{*} & -      & -     & -      & 
80.5 & -      & - \\
LSTM (30Hz)~\cite{Dipietro:MICCAI16}\text{*} & -      & -     & -      & 
78.38 & -      & - \\
BiLSTM (5Hz)~\cite{Dipietro:MICCAI16}\text{*} & -      & -     & -      & 
\emph{83.3} & -      & - \\
BiLSTM (30Hz) ~\cite{Dipietro:MICCAI16}\text{*} & -      & -     & -      & 
80.15 & -      & - \\
TCN~\cite{Lea:ECCV16-WBNIMR} & -      & -     & -      & 
79.6 & -      & - \\
LC-SC-CRF~\cite{Lea:ICRA16}\text{**} & -      & -     & -      & 
\textbf{83.4} & -      & - \\
Ours & \emph{86.21} (0.34) & \emph{83.89} (0.08) & \textbf{75.19} (0.12) & 78.16 (0.42) &\emph{76.68} (1.20) & \emph{66.25} (0.06)\\
\hline
\end{tabular}
\caption{Average per-frame action recognition accuracy for surgical task segmentation and recognition on the JIGSAWS dataset~\cite{Gao2014}. The results are averaged over three random runs, with the standard deviation reported in parentheses. Best results are shown in bold, while second best results are denoted in italics.\text{*} Our results are not directly comparable with those of~\cite{Dipietro:MICCAI16}, since they were using data downsampled in time (5Hz). For a fair comparison, results for LSTM, BiLSTM on non-downsampled data (30Hz) were obtained using the code and default parameters publicly available from the authors~\cite{Dipietro:MICCAI16}. \text{**} Our results are not directly comparable with those of LC-SC-CRF~\cite{Lea:ICRA16}, where authors were using both kinematic data as well as the distance from the tools to the closest object in the scene from the video. }
\vspace{-1em}
\label{tab:jigsaws_soa}
\end{table*}
We evaluate our method on two public datasets for fine-grained action segmentation and recognition: JIGSAWS~\cite{Gao2014} and 50 Salads~\cite{Stein:UbiComp13}. First, we report our results on each
dataset and compare them with the state of the art. Next, we
examine the effect of different model components.

\subsection{Datasets}
\myparagraph{JHU-ISI Gesture and Skill Assessment (JIGSAWS)~\cite{Gao2014}} This dataset provides kinematic data of the right and left manipulators of the master and slave da Vinci surgical robot recorded at $30$ Hz during the execution of three surgical tasks (Suturing (SU), Knot-tying (KT) and Needle-passing (NP)) by surgeons with varying skill levels. In particular, kinematic data include positions, orientations, velocities etc. ($76$ variables in total), and there are 8 surgeons performing a total of 39, 36 and 26 trials for the Suturing, Knot-tying and Needle-passing surgical tasks, respectively. This dataset is challenging due to the significant variability in the execution of tasks by surgeons of different skill levels and the subtle differences between fine-grained actions. There are 10, 6 and 8 different action classes for the Suturing, Knot-tying and Needle-passing tasks, respectively. Examples of action classes are \emph{orienting needle}, \emph{reaching for needle with right hand}, \emph{pulling suture with left hand}, and \emph{making C loop}. We evaluate our method using the standard Leave-One-User-Out (LOUO) and Leave-One-Supertrial-Out (LOSO) cross-validation setups~\cite{Ahmidi:TBME17}.

\myparagraph{50 Salads~\cite{Stein:UbiComp13}} This dataset provides data recorded by 10 accelerometers attached to kitchen tools, such as knife, peeler, oil bottle etc., during the preparation of a salad by 25 users. This dataset features annotations at four levels of granularity, out of which we use the \emph{eval} and \emph{mid} granularities. The former consists of 10 actions that can be reasonably recognized based on the utilization of accelerometer-equipped objects, such as \emph{add oil}, \emph{cut}, \emph{peel} etc., while the latter consists of 18 mid-level actions, such as \emph{cut tomato}, \emph{peel cucumber}. Both granularities include a background class. We evaluate our method using the ground truth labels and the 5-fold cross-validation setup proposed by the authors of~\cite{Lea:ECCV16-WBNIMR,Lea:CVPR17}. 

In summary, these two datasets provide kinematic/sensor data recorded during the execution of long goal-driven complex activities, which are comprised of multiple fine-grained action instances following a grammar. Hence, they are suitable for evaluating our method, which was designed for kinematic data and features a temporal model that is able to capture action transitions. Other datasets collected for action segmentation with available skeleton data, such as CAD-120~\cite{Koppula:IJRR13}, Composable Activities~\cite{Lillo:CVPR14}, Watch-n-Patch~\cite{Wu:CVPR15} and OAD~\cite{De:ECCV16}, have a mean number of 3 to 12 action instances per sequence~\cite{Liu:CoRR17}, while for example the Suturing task in the JIGSAWS dataset features an average of 20 action instances per sequence, ranging from 17 to 37. It is therefore more challenging for comparing temporal models. Recently, the PKU-MMD dataset~\cite{Liu:CoRR17} was proposed, which is of larger scale and also contains around 20 action instances per sequence. However, the actions in this dataset are not fine-grained (e.g., \emph{hand waving}, \emph{hugging} etc.).
\subsection{Implementation Details}
Input data are normalized to have zero mean and unit standard deviation. We apply PCA on the robot kinematic data of the JIGSAWS dataset to reduce their dimension from $76$ to $35$ following the setup of~\cite{Sefati:M2CAI15}. The dictionary is initialized using the SPAMS dictionary learning toolbox~\cite{Mairal:JMLR2010} and the CRF parameters are initialized to $0$. We use Stochastic Gradient Descent with a batch size of $1$ and momentum of $0.9$. We also reduce the learning rate by one half every $20$ epochs and train our models for $100$ epochs. Parameters such as the regularization cost $C$, initial learning rate $\eta$, temporal window size for average-pooling $L$, Lasso regularizer parameter $\lambda_u$, skip chain length $d$ and dictionary size $m$ vary with each dataset, surgical task or granularity. The window size was fixed to $71$ for JIGSAWS and $51$ for 50 Salads, the dictionary size $M$ was chosen via cross-validation from the values $\{50, 100, 150, 200\}$, $\lambda_u$ from values $\{0.1, 0.5\}$, $C$ from $\{0.001, 0.01, 0.1, 1\}$, $\eta$ from $\{0.0001, 0.001, 0.01\}$ and $d$ from $\{21, 51, 81\}$. To perform cross-validation we generate five random splits of the available sequences of each dataset task/granularity. Note that since both datasets have a fixed test setup, with all users appearing in the test set exactly once, it is not clear how to use them for hyperparameter selection without inadvertently training on the test set. Here we randomly crop a temporal segment from each of the videos instead of using the whole sequences for cross-validation, in order to avoid using the exact same video sequences which will be used for evaluating our method. The length of these segments is $80\%$ of the original sequence length. Furthermore, we select $m$, $\lambda_u$ and $d$ by using the initialized dictionary $\boldsymbol {\Psi}^0$ and learning the weights of a SC-CRF, while we choose $C$ and $\eta$ by jointly learning the dictionary and the SC-CRF weights. 
\subsection{Results}

\myparagraph{Overall performance} We first compare our method with state-of-the-art methods on the JIGSAWS and 50 Salads datasets. The per-frame action recognition accuracies of all the compared methods on JIGSAWS are summarized in Table~\ref{tab:jigsaws_soa}. It can be seen that our method yields the best or second best performance for all tasks on both the LOSO and LOUO setups, except for Suturing LOUO, where  LC-SC-CRF achieves per-frame action recognition accuracies up to $83\%$. However, their result is not directly comparable to ours, since they employ additional video-based features. Also note that in~\cite{Lea:WACV15} they use a SC-CRF with an additional pairwise term (skip-length data potentials), which is not incorporated in our model and could potentially improve our results. However, it is worth noting that our method achieves comparable performance to deep recurrent models such as LSTMs~\cite{Dipietro:MICCAI16} and the newly proposed TCN~\cite{Lea:ECCV16-WBNIMR},  which possibly captures complex temporal patterns, such as action compositions, action durations, and long-range temporal dependencies. Furthermore, our method consistently improves over SDSDL~\cite{Sefati:M2CAI15}, which was based on joint sparse dictionary and linear SVM learning, as well as a temporal smoothing of results using the Viterbi algorithm with precomputed action transition probabilities. 

Table~\ref{tab:fiftysalads_sota} summarizes our results on the 50 Salads dataset under two granularities. Although the modality used in this dataset is different (accelerometer data), it can be seen that our method is very competitive among all the compared methods, even with respect to methods relying on powerful deep temporal models such as LSTMs. 

\begin{table}[t]
\centering
\begin{tabular}{|@{\,}c@{\,}|@{\,}c@{\,}|@{\,}c@{\,}|}
\hline
Method & \multicolumn{2}{c|}{50 Salads} \\
\hline
 & \emph{eval} & \emph{mid} \\
\hline

LC-SC-CRF~\cite{Lea:ICRA16} & 77.8 & \emph{55.05}\text{*} \\
LSTM~\cite{Lea:ECCV16-WBNIMR} & 73.3 & - \\
TCN~\cite{Lea:ECCV16-WBNIMR} & \textbf{82.0} & - \\
Ours & \emph{80.04} (0.11) & \textbf{56.72} (0.72) \\
\hline 
\end{tabular}
\caption{Results for action segmentation and recognition on the 50 Salads dataset using granularities \emph{eval} and \emph{mid}. Results are averaged over three random runs, with the standard deviation reported in parentheses. Best results are shown in bold, while second best results are denoted in italics.\text{*} LC-SC-CRF~\cite{Lea:ICRA16} was evaluated on the \emph{mid} granularity with smoothed out short interstitial background segments~\cite{Lea:ECCV16-WBNIMR}. 
}
\vspace{-1.5em}
\label{tab:fiftysalads_sota}
\end{table}

\myparagraph{Ablative analysis} 
In Tables~\ref{tab:jigsaws_ablation},~\ref{tab:fiftysalads_ablation} we analyze the contribution of the key components of our method, namely the contribution of
a) using sparse features (Eq.~\ref{eq:zs}) obtained from an unsupervised dictionary in conjunction with a Linear Chain CRF, b) substituting the Linear Chain CRF with a Skip Chain CRF (SC-CRF) and c) jointly learning the dictionary used in sparse coding and the CRF unary and pairwise weights. As expected, using sparse features instead of the raw kinematic features consistently boosts performance across all tasks on JIGSAWS. Similarly, sparse coding of accelerometer data improves performance on 50 Salads and notably this improvement is larger in the case of fine-grained activities (\emph{mid} granularity). Furthermore, using a SC-CRF further boosts performance as expected, since it is more suitable for capturing action-to-action transition probabilities in contrast to the Linear Chain CRF which captures frame-to-frame action transition probabilities. 

It is however surprising that learning a discriminative dictionary jointly with the CRF weights does not significantly improve performance, yielding an improvement of at most $\sim 1\%$. Further investigating this result, we computed additional metrics for evaluating the segmentation quality on the JIGSAWS dataset. In particular, we report the edit score~\cite{Lea:ICRA16}, a metric measuring how well the model predictions the ordering of action segments, and segmental-f1@10 score as defined in~\cite{Lea:CVPR17}. As it can be seen in Table~\ref{tab:jigsaws_additional_metrics}, performance is similar across all metrics for both unsupervised and discriminative dictionary, except for a consistent improvement in Needle Passing. One possible explanation could be that the computation of features based on average pooling of sparse codes in a temporal window might reduce the impact of the discriminatively trained dictionary. However, repeating the experiment on JIGSAWS (Suturing LOSO) without average temporal pooling leads to the same behavior, i.e. using a dictionary learned via unsupervised training with a SC-CRF yields a per-frame accuracy of $86.64\%$, while using a dictionary jointly trained with the SC-CRF yields $86.11\%$. Our findings could be attributed to the limited training data. They also seem to corroborate the conclusions drawn by Coates et al.~\cite{Coates:ICML11-sparse}, who have experimentally observed that the superior performance of sparse coding, especially when training samples are limited, arises from its non-linear encoding scheme and not from the basis functions that it uses. 
\begin{table}[t]
\centering
\begin{tabular}{|@{\,}l@{\,}|@{\,}c@{\,}|@{\,}c@{\,}|}
\hline
Method & \multicolumn{2}{c|}{50 Salads} \\
\hline
 & \emph{eval} & \emph{mid} \\
\hline
raw + CRF & 71.81 (0.55) & 44.83 (0.73) \\
SF + CRF & 76.65 (0.19) & 52.63 (0.23)\\
SF + SC-CRF & 80.24 (0.20) & \textbf{56.73} (0.08)\\
SDL + SC-CRF & \textbf{80.54} (0.11) & 56.72 (0.72)\\
\hline 
\end{tabular}
\caption{Analysis of contribution to recognition performance from each
model component in the 50 Salads dataset. Results are averaged over three random runs, with the standard deviation reported in parentheses.  raw+CRF: use kinematic data as input to a CRF, SF + CRF: use sparse features $z$ as input to a CRF, SF + SC-CRF: use sparse features $z$ as input to a SC-CRF, SDL + SC-CRF: joint dictionary and SC-CRF learning.}
\vspace{-1.5em}
\label{tab:fiftysalads_ablation}
\end{table}

\begin{table*}[ht!]
\setlength{\tabcolsep}{0.75mm}
\centering
\begin{tabular}{|@{\;}l@{\;}|ccc|ccc|}
\hline
Method & \multicolumn{3}{|c|}{LOSO} & \multicolumn{3}{|c|}{LOUO} \\
\hline
 & SU & KT & NP & SU & KT & NP\\
\hline
raw + CRF & 79.57 (0.04) & 76.39 (0.09) & 66.24 (0.10) & 71.77 (0.05) & 69.63 (0.06) & 59.47 (0.18) \\
SF + CRF & 85.70 (0.01)& 82.06 (0.03) & 71.72 (0.07) & 76.64 (0.05) & 73.58 (0.07) & 60.59 (0.19) \\
SF + SC-CRF & \textbf{87.60} (0.03) & 83.71 (0.03) & 74.63 (0.02) & \textbf{79.95} (0.05) & \textbf{76.88} (0.14) & 65.75 (0.12) \\
SDL + SC-CRF & 86.21 (0.34) & \textbf{83.89} (0.07) & \textbf{75.19} (0.12) & 78.16 (0.42) & 76.68 (1.20) & \textbf{66.25} (0.06) \\
\hline 
\end{tabular}
\caption{Analysis of contribution to recognition performance from each
model component in the JIGSAWS dataset. Results are averaged over three random runs, with the standard deviation reported in parentheses.  raw+CRF: use kinematic data as input to a Linear Chain CRF, SF + CRF: use sparse features $z$ as input to a CRF, SF + SC-CRF: use sparse features $z$ as input to a SC-CRF, SDL + SC-CRF: joint dictionary and SC-CRF learning.}
\label{tab:jigsaws_ablation}
\end{table*}

\begin{table*}
\setlength{\tabcolsep}{0.75mm}
\small
\centering
\begin{tabular}{|@{\;}l@{\;}|ccc|ccc|}
\hline
Method & \multicolumn{3}{|c|}{LOSO} & \multicolumn{3}{|c|}{LOUO} \\
\hline
 & SU & KT & NP & SU & KT & NP\\
\hline
SF + SC-CRF & \textbf{87.57}/\textbf{82.92}/\textbf{88.59} & 83.08/\textbf{82.87}/87.46 & 74.62/73.05/76.01 & \textbf{79.92}/\textbf{63.39}/\textbf{75.00} & \textbf{76.93}/63.61/71.38 & 65.81/55.45/62.30 \\
SDL + SC-CRF & 85.90/75.45/83.47 & \textbf{83.97}/82.82/\textbf{87.94} & \textbf{75.33}/\textbf{76.63}/\textbf{79.85} & 78.42/58.02/69.22 & 76.39/\textbf{65.55}/\textbf{72.87} & \textbf{66.29}/\textbf{60.85}/\textbf{64.43}  \\
\hline
\end{tabular}
\caption{Comparison of unsupervised and supervised dictionary used for sparse coding on JIGSAWS dataset. Metrics reported are: accuracy/edit score/segmental f1 score. Results are from a single random run. SF + SC-CRF: use sparse features $z$ obtained from unsupervised dictionary as input to a SC-CRF, SDL + SC-CRF: use sparse features $z$ from discriminative dictionary learned jointly with a SC-CRF.} 
\label{tab:jigsaws_additional_metrics}
\end{table*}

\begin{figure*}[ht!]
\centering
\begin{tabular}{@{}cc@{}}
\subcaptionbox{Suturing LOSO}{
 \begin{tabular}{@{}c@{}}
   \includegraphics[trim={0 0 0 0},clip,scale=0.2]{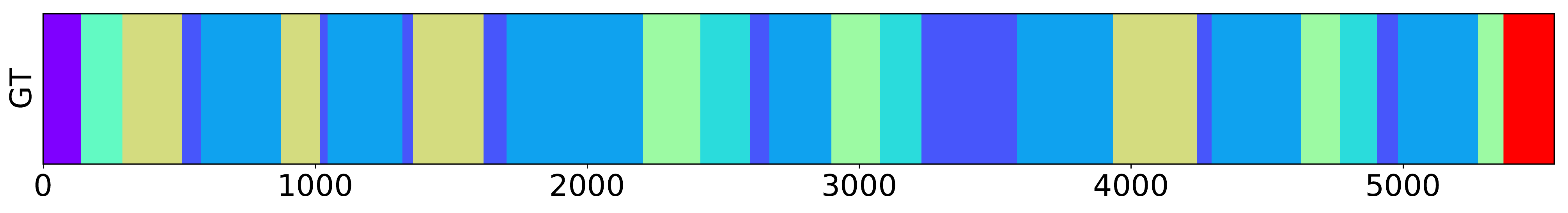} \\
\includegraphics[trim={0 0 0 0},clip,scale=0.2]{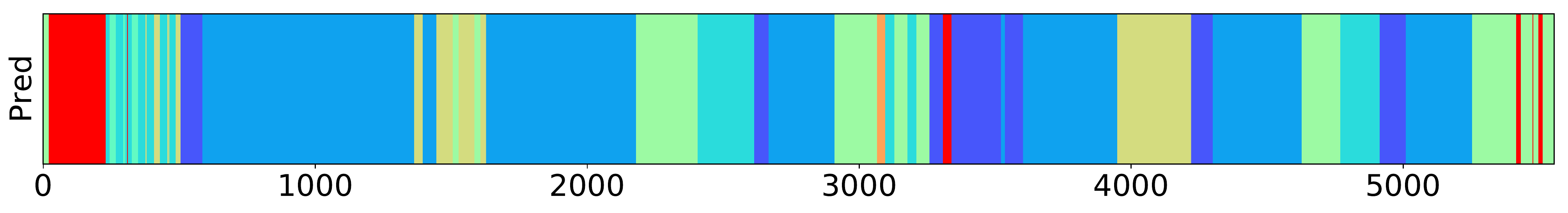} \\
\includegraphics[trim={0 0 0 0},clip,scale=0.2]{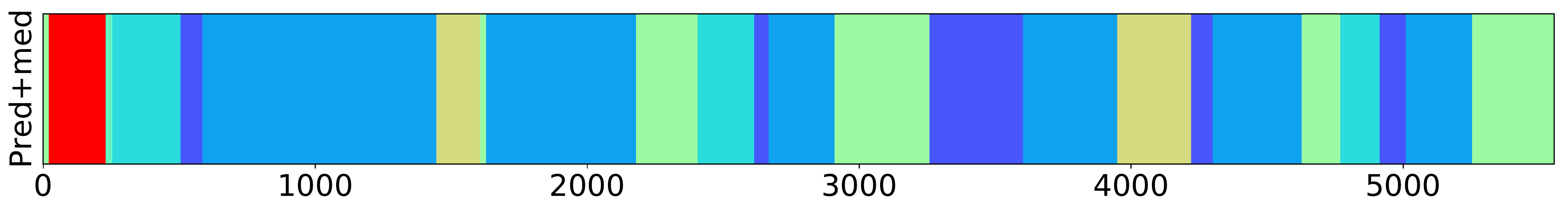} \\
  \end{tabular}
} &

\subcaptionbox{Suturing LOSO}{
 \begin{tabular}{@{}c@{}}
\includegraphics[trim={0 0 0 0},clip,scale=0.2]{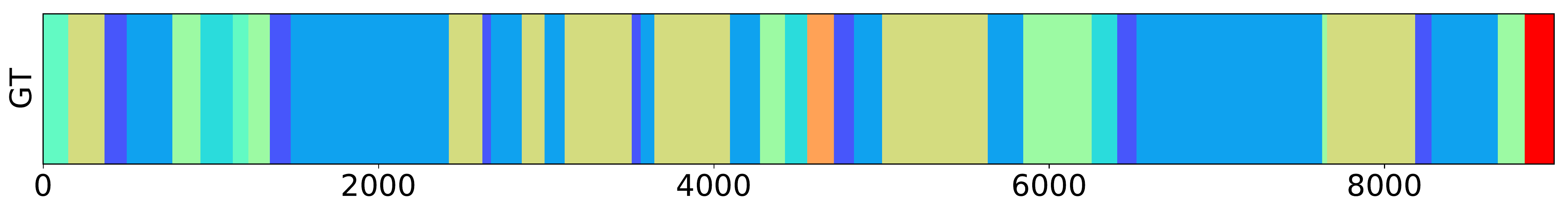}\\
\includegraphics[trim={0 0 0 0},clip,scale=0.2]{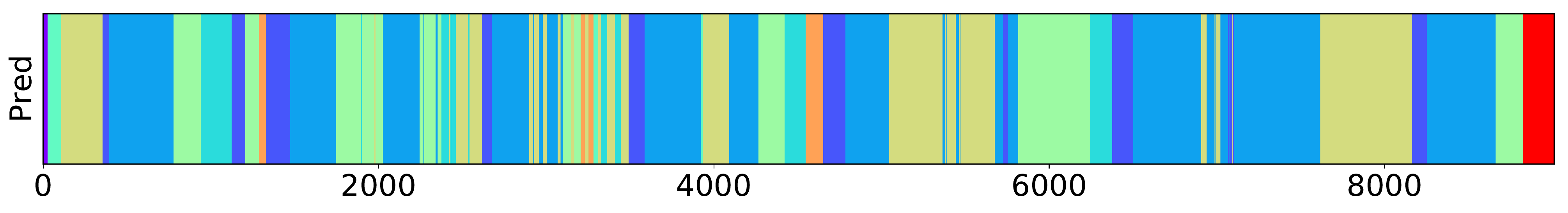}\\
\includegraphics[trim={0 0 0 0},clip,scale=0.2]{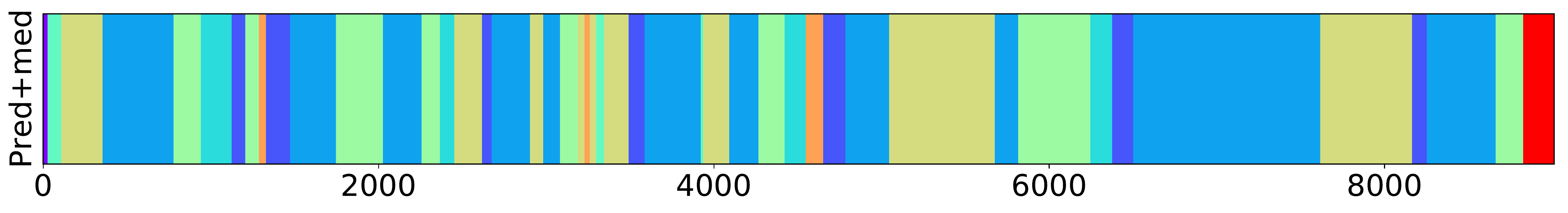} \\
  \end{tabular}
} \\

\subcaptionbox{Suturing LOUO}{
 \begin{tabular}{@{}c@{}}
\includegraphics[trim={0 0 0 0},clip,scale=0.2]{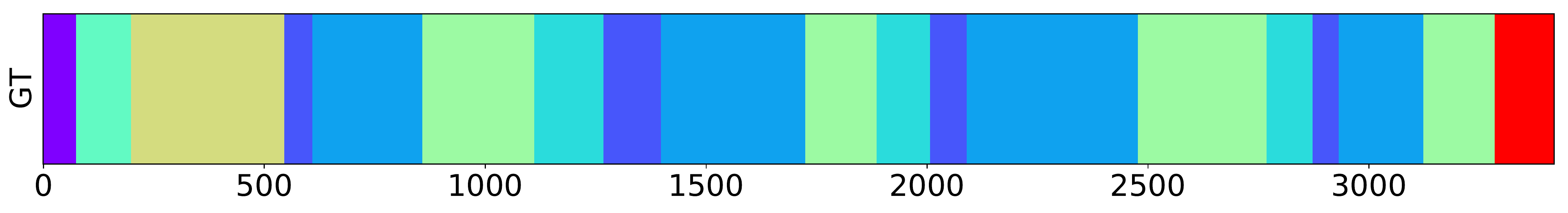}\\
\includegraphics[trim={0 0 0 0},clip,scale=0.2]{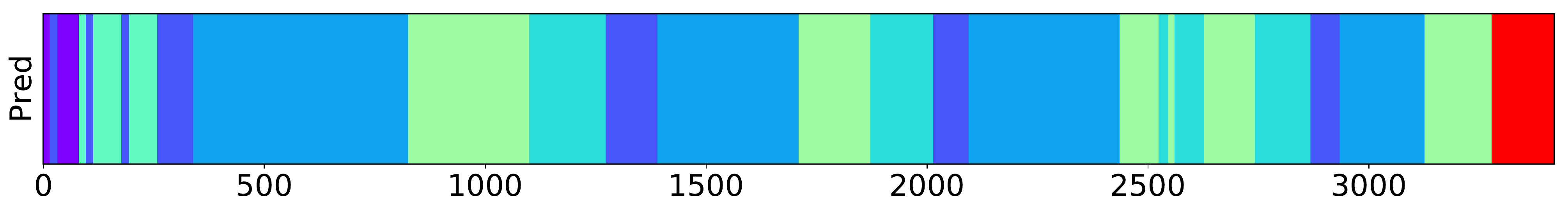}\\
\includegraphics[trim={0 0 0 0},clip,scale=0.2]{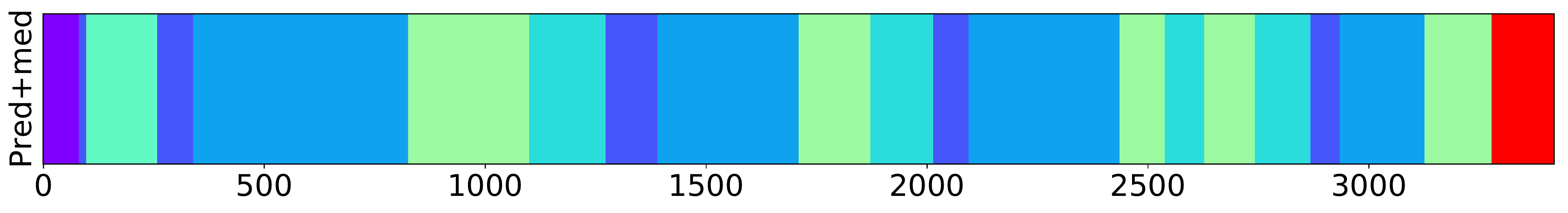}\\
  \end{tabular}
} &

\subcaptionbox{Suturing LOUO}{
 \begin{tabular}{@{}c@{}}
\includegraphics[trim={0 0 0 0},clip,scale=0.2]{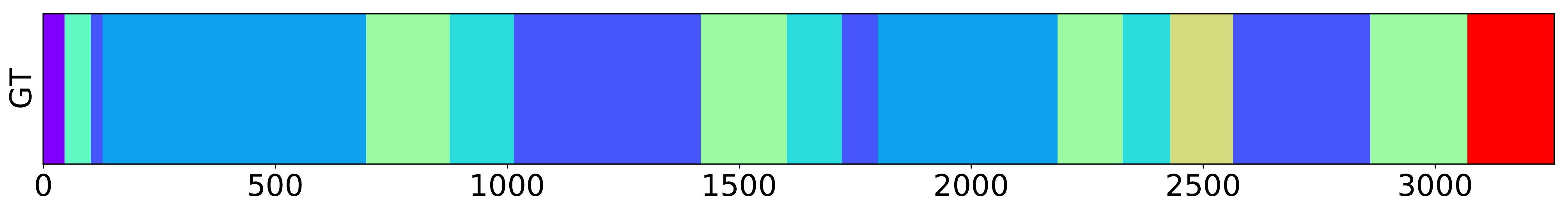}\\
\includegraphics[trim={0 0 0 0},clip,scale=0.2]{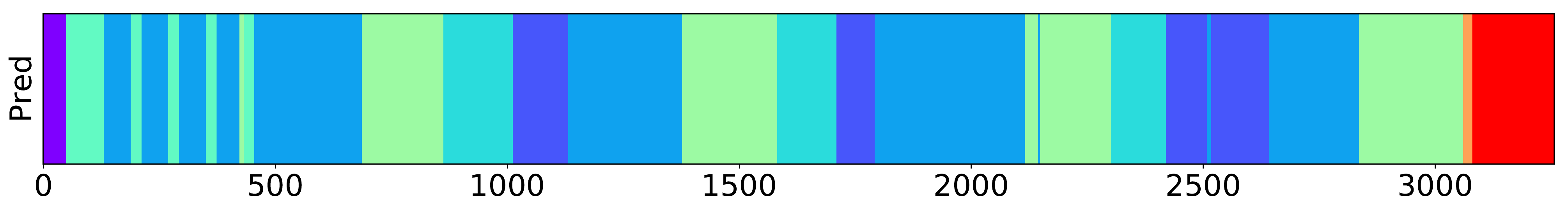}\\
\includegraphics[trim={0 0 0 0},clip,scale=0.2]{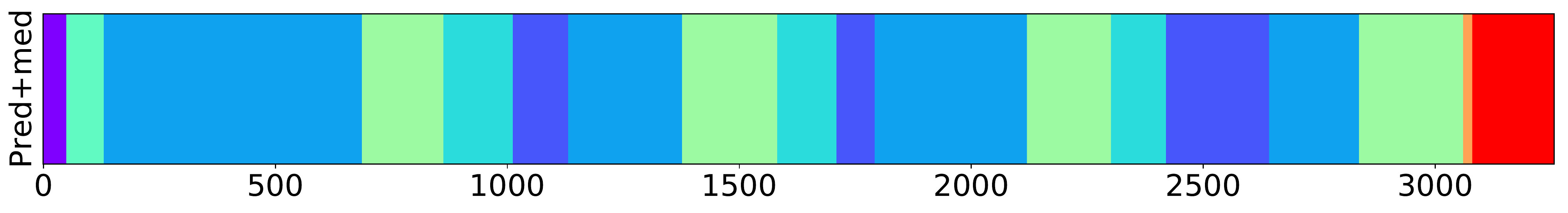} \\
  \end{tabular}
} \\
\end{tabular}
   \caption{Qualitative examples of ground truth and predicted temporal segmentations (before and after median filtering) on JIGSAWS data.  Each color denotes a different action class. (Best viewed in color.)}
   \vspace{-1em}
\label{fig:segmentations}
\end{figure*}

\myparagraph{Qualitative results} In Fig.~\ref{fig:segmentations} we show examples of ground truth segmentations and predictions for selected testing sequences from JIGSAWS Suturing. As it can be seen, the LOUO setup is more challenging since the model is asked to recognize actions performed by a user it has not seen before and in addition to that there is great variability in experience and styles between surgeons. In all cases our model outputs smooth predictions, without significant over-segmentations.
\vspace{-2mm}
\section{Conclusion}
We have presented a novel end-to-end learning framework for fine-grained action segmentation and recognition, which combines features based on sparse coding with a Linear Chain CRF model. We also proposed a max-margin approach for jointly learning the sparse dictionary and the CRF weights, resulting in a dictionary adapted to the task of action segmentation and recognition. Experimental evaluation of our method on two datasets showed that our method performs on par or outperforms most of the state-of-the-art methods. 
Given the recent success of deep convolutional networks (CNNs), future work will explore using deep features as inputs to the temporal model and jointly learning the CNN and CRF parameters in a unified framework.

\myparagraph{Acknowledgements}
We would like to thank Colin Lea and Lingling Tao for their insightful comments and for their help with the JIGSAWS dataset, and Vicente Ord\'o\~nez for his useful feedback during this research collaboration. This work was supported by NIH grant R01HD87133.

{\small
\bibliographystyle{ieee}
\bibliography{biblio/vidal,biblio/vision,biblio/recognition,biblio/learning,biblio/sparse,biblio/segmentation,biblio/dataset,biblio/surgery}
}
\end{document}